%% file: sample-sigconf.tex
\documentclass[sigconf]{acmart}

\AtBeginDocument{%
  }

\copyrightyear{2024}
\acmYear{2024}
\setcopyright{acmlicensed}\acmConference[MM '24]{Proceedings of the 32nd ACM International Conference on Multimedia}{October 28-November 1, 2024}{Melbourne, VIC, Australia}
\acmBooktitle{Proceedings of the 32nd ACM International Conference on Multimedia (MM '24), October 28-November 1, 2024, Melbourne, VIC, Australia}
\acmDOI{10.1145/3664647.3680982}
\acmISBN{979-8-4007-0686-8/24/10}

\usepackage{hyperref}       
\usepackage{url}            
\usepackage{booktabs}       
\usepackage{amsfonts}       
\usepackage{nicefrac}       
\usepackage{microtype}      
\usepackage{xcolor}         
\usepackage{graphicx}
\usepackage{amsmath}
\usepackage{multirow, multicol, colortbl, float}
\usepackage{wrapfig}
\usepackage{tcolorbox}
\usepackage{subfigure}
\usepackage{caption}
\usepackage{subcaption}
\usepackage{enumitem,kantlipsum}
\newcommand{\VarSty}[1]{\textnormal{\ttfamily\color{blue!90!black}#1}\unskip}

\begin{document}

\title{PTSBench: A Comprehensive Post-Training Sparsity Benchmark Towards Algorithms and Models}

\author{Zining Wang}
\affiliation{%
  \institution{SKLCCSE, Beihang University}
  \city{Beijing}
  \country{China}
}
\email{wangzining@buaa.edu.cn}
\orcid{0000-0002-1259-5377}

\author{Jinyang Guo}
\affiliation{%
  \institution{SKLCCSE, Beihang University}
  \city{Beijing}
  \country{China}
}
\email{jinyangguo@buaa.edu.cn}
\orcid{0000-0003-1956-3367}
\authornote{Corresponding author.}

\author{Ruihao Gong}
\affiliation{%
  \institution{SKLCCSE, Beihang University}
  \city{Beijing}
  \country{China}
}
\affiliation{%
   \institution{SenseTime Research}
  \city{Beijing}
  \country{China}
}
\email{gongruihao@sensetime.com}
\orcid{0000-0002-6024-7086}

\author{Yang Yong}
\affiliation{%
 \institution{SenseTime Research}
  \city{Shanghai}
  \country{China}
  }
 \email{yongyang@sensetime.com}
 \orcid{0009-0004-2804-7801}

\author{Aishan Liu}

\affiliation{%
  \institution{SKLCCSE, Beihang University}
  \city{Beijing}
  \country{China}
  }
  \affiliation{%
  \institution{Institute of Dataspace}
  \city{Hefei}
  \country{China}
  }
  \email{liuaishan@buaa.edu.cn}
  \orcid{0000-0002-4224-1318}

\author{Yushi Huang}
\affiliation{%
  \institution{Beihang University}
  \city{Beijing}
  \country{China}
  }
  \affiliation{%
  \institution{SenseTime Research}
  \city{Beijing}
  \country{China}
  }
\email{huangyushi@sensetime.com}
\orcid{0009-0002-7898-8402}

\author{Jiaheng Liu}
\affiliation{%
  \institution{Nanjing University}
  \city{Nanjing}
  \country{China}
  }
\email{buaaljiaheng@gmail.com}
\orcid{0000-0002-5183-8538}


\author{Xianglong Liu}
\affiliation{%
  \institution{SKLCCSE, Beihang University}
  \city{Beijing}
  \country{China}
  }
  \affiliation{%
  \institution{Institute of Dataspace}
  \city{Hefei}
  \country{China}
  }
\email{xlliu@buaa.edu.cn}
\orcid{0000-0001-8425-4195}
\renewcommand{\shortauthors}{Zining Wang et al.}

\input{sec/abstract}




\maketitle

\input{sec/introduction}

\input{sec/background}
\input{sec/overview}
\input{sec/implementation}

\input{sec/experiment}
\input{sec/discussion}

\begin{acks}
This work was supported by the National Science and Technology Major Project (2021ZD0110503), Beijing Municipal Science and Technology Project (Nos. Z231100010323002),  and the National Natural Science Foundation of China (No. 62306025, No. 92367204).
\end{acks}

\input{sample-sigconf.bbl}

\bibliographystyle{ACM-Reference-Format}
\bibliography{sample-sigconf}


\end{document}

%% file: sec/abstract.tex
\begin{abstract}
With the increased attention to model efficiency, 
post-training sparsity (PTS) has become more and more prevalent because of its effectiveness and efficiency. However, there remain questions on better practice of PTS algorithms and the sparsification ability of models, which hinders the further development of this area. 
Therefore, a benchmark to comprehensively investigate the issues above is urgently needed. 
In this paper, we propose the first comprehensive post-training sparsity benchmark called PTSBench towards algorithms and models. 
We benchmark 10+ PTS general-pluggable fine-grained techniques on 3 typical tasks using over 40 off-the-shelf model architectures. 
Through extensive experiments and analyses, we obtain valuable conclusions and provide several insights from both algorithms and model aspects.
Our PTSBench can provide (1) new observations for a better understanding of the PTS algorithms, (2) in-depth and comprehensive evaluations for the sparsification ability of models, 
and (3) a well-structured and easy-integrate open-source framework.
We hope this work will provide illuminating conclusions and advice for future studies of post-training sparsity methods and sparsification-friendly model design. The code for our PTSBench is released at \href{https://github.com/ModelTC/msbench}{https://github.com/ModelTC/msbench}.

\end{abstract} 

%% file: sec/introduction.tex
\section{Introduction}
\label{sec:introduction}


Although deep learning has been widely used in various fields, it requires a considerable amount of memory and computational power. To address this issue, many strategies have emerged to compress the model, including model quantization~\cite{krishnamoorthi2018quantizing, huang2024tfmq,jacob2018quantization,hubara2017quantized,nagel2019data, lv2024ptq4sam, gong2024llm}, model sparsification~\cite{guo2020multi,wang2020picking,he2018amc,hassibi1992second,li2022revisiting, frantar2022optimal, guo2024compressing}, network distillation~\cite{hinton2015distilling, guo2023adaptive, guo2020channel}, lightweight network design~\cite{sandler2018mobilenetv2} and weight matrix decomposition~\cite{dziugaite2015neural}. One of the most representative methods is model sparsification, which involves removing unimportant weights from the model. Among all the sparsification methods, post-training sparsity (PTS) has received much attention in recent years because of its small training cost.

In the scenario of post-training sparsity (PTS), we are given a pre-trained dense model along with a small amount of unlabeled calibration data. We aim to generate an accurate sparse model without an end-to-end retraining process.
Under these settings, several representative methods have been proposed, including POT \cite{lazarevich2021post}, AdaPrune \cite{hubara2021accelerated}, and OBC \cite{frantar2022optimal}. These state-of-the-art (SOTA) methods have achieved almost no performance loss after sparsification.
However, even though the high-performance PTS methods have reached, there are still two problems remain:

\textbf{\textit{Problem-1:}} \textbf{Post-training sparsity algorithm exploration is incomplete.}
Current PTS methods~\cite{lazarevich2021post,hubara2021accelerated,frantar2022optimal} share the same sparsification paradigm: they first allocate sparsity rate to each layer to sparsify the model and then reconstruct the activation to recover the performance further.
However, while current research follows this pipeline, it still lacks a fine-grained exploration of PTS techniques. For example, most PTS methods adopt layer-wise reconstruction granularity in the reconstruction process. However, block-wise reconstruction granularity has been proven effective in quantization approaches~\cite{li2021brecq, wei2022qdrop} but is not explored in PTS algorithms. The absence of in-depth analysis of fine-grained techniques hinders further development of PTS approaches. Thus, benchmarking toward fine-grained techniques in PTS algorithms is urgently required.

\textbf{\textit{Problem-2:}} \textbf{Relationship between model and sparsity remains unexplored.} 
Current PTS research heuristically chooses commonly used models (e.g., ResNet and RegNetX) or datasets to validate their methods. However, it still lacks a comprehensive evaluation of the relationship between models and sparsity. In real-world applications, we often face the scenario that multiple network architectures with similar sizes can be used as the backbone, and we need to sparsify them before deployment for efficient inference. It is still an open question about which network architecture is sparsity-friendly for better performance. Moreover, as the deployment platform varies, we also need to use different model sizes (e.g., different layer numbers). It is unclear whether a network architecture with different sizes is robust for the sparsity algorithms. 
In addition, model structures typically require various designs and modifications tailored to each task under different tasks. The extent to which these task-oriented designs and modifications are conducive to sparsity is also unknown.
Therefore, conducting a comprehensive evaluation from an architectural perspective is also necessary from both theoretical and practical aspects. 


To address the problems mentioned above, in this paper, we present \textbf{PTSBench}, a \textbf{P}ost-\textbf{T}raining \textbf{S}parsification \textbf{Bench}mark to evaluate the PTS technique from both algorithm and model aspects comprehensively.
Starting from the real-world model production requirements, we carefully design 5 tracks for comparison. With over 8000 A800 GPU hours consumption, we benchmark 40+ classical off-the-shelf models, 3 typical computer vision tasks, and 10+ easy-pluggable fine-grained techniques in post-training sparsity. Based on the evaluation, we provide in-depth analyses of PTS methods from both algorithm and model perspectives and offer useful insights and guidance on PTS method design and validation.

Overall, our contributions can be summarized as follows:
\begin{enumerate}
    \item \textbf{Comprehensive benchmark.} We construct PTSBench, which is the first systematic benchmark to conduct a comprehensive evaluation of PTS methods. 
    It provides a brand new perspective to evaluate post-training sparsity from both algorithm and model aspects.
    \item \textbf{In-depth analysis.} Based on extensive experiments, we uncover and summarize several useful insights and take-away conclusions, which can serve as a guidance for future PTS method design.
    \item \textbf{Well-structured open-source framework.} We have released our open-source benchmark codes repository. Research communities can easily use our platform to evaluate sparsity approaches. It can also serve as a well-organized codebase for future research of PTS algorithms.
    
\end{enumerate}


%% file: sec/background.tex
\section{Background}
\label{sec:background}

\begin{figure*}[t]
  \centering
  \includegraphics[width=0.9\linewidth]{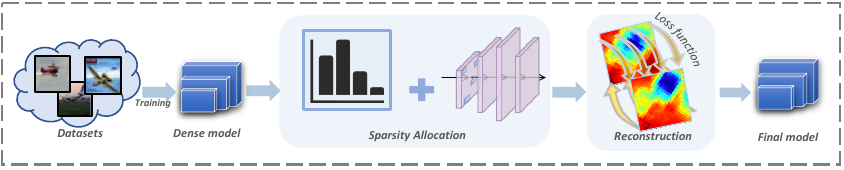 }
  \caption{The illustration of the overall Post-Training Sparsity pipeline, which is employed by most PTS methods.}
  \label{fig:pipeline}
\end{figure*}

\subsection{Post-training Sparsity}
\label{sec:background:pipeline}

Post-Training Sparsity (PTS) aims to sparsify a pre-trained neural network while preserving 
its accuracy on a specific task. All of the current PTS researches~\cite{lazarevich2021post, hubara2021accelerated, frantar2022optimal, gong2024fast} adopt two-step sparsity paradigm including sparsity allocation and reconstruction, as shown in Figure \ref{fig:pipeline}. 
In the sparsity allocation procedure, a specific sparsity rate is allocated to each layer following a predetermined metric, and weights at corresponding positions are zeroed based on sparsity criteria. The reconstruction process involves employing a series of techniques to recover model accuracy drop from sparsification. We will give a detailed introduction to these two parts. 

\subsection{Sparsity Allocation}
Many previous works have shown that allocating a more reasonable sparsity for each layer can lead to a more effective sparsification outcome \cite{evci2020rigging, han2015learning, he2018amc, liu2017learning, gong2024fast}. 
Currently, sparsity allocation methods can be categorized into the following three types.




\begin{enumerate}
    \item \textbf{Heuristic based strategy.} In this type, the sparsity ratio for each layer is predetermined manually, such as uniform sparsity\cite{mostafa2019parameter, gale2019state}. 
    \item \textbf{Criterion based strategy.} Weights across all layers are ranked according to a specific metric, and a certain percentage of the weights with lower scores are set as zero. The corresponding sparsity rates for each layer can be naturally obtained\cite{han2015learning, evci2020rigging, renda2020comparing, frankle2019stabilizing}.
    \item\textbf{Learning based strategy.} This strategy learns sparsity rates for each layer by optimizing a loss function to achieve optimal sparsity allocation \cite{gong2024fast, kusupati2020soft}.
\end{enumerate}

Although current work proposed multiple sparsity rate allocation strategies, 
these works lack evaluation on broader architectures, sizes, and tasks. Moreover, while there is a strong focus on the effectiveness of the methods, it still lacks in-depth analysis, such as why allocating a sparsity rate in a certain way can reach high performance. These two limitations in current research pose the question of better practice for PTS algorithms.

\subsection{Reconstruction}
\label{sec:reconstruction}





After sparsity allocation, PTS methods will apply reconstruction to reconstruct the sparse activation for compensating the accuracy loss caused by sparsity. 
In this paper, we benchmark three fine-grained pluggable techniques that are often identified as influencing the effectiveness of sparsity in this process: error correction, reconstruction input, and reconstruction granularity.

\subsubsection{Error Correction}
Error correction is widely used in many post-training quantization (PTQ)~\cite{nagel2019data, nagel2021white, he2023ptqd} methods. It aims to align the weight distribution after compression with the original weight distribution. However, current PTS methods do not comprehensively and systematically evaluate this procedure. Specifically, the error correction procedure can be written as follows:


\begin{equation}
\begin{aligned}
  &\hat{\textbf{W}_s} = \lambda \textbf{W}_s+E(\textbf{W}_{d})-E(\lambda \textbf{W}_s), \\
  \text{and }\hat{b_s} &= b_d + E(f(\textbf{W}_d,\textbf{X}_d))-E(f(\hat{\textbf{W}_s},\textbf{X}_d)), \\
  &\text{where } \lambda =\frac{\sigma(\textbf{W}_d)}{\sigma(\textbf{W}_s)+\epsilon}.
  \label{eq:ec1}
\end{aligned}
\end{equation}
$\hat{\textbf{W}_s}$ and $\hat{b_s}$ are the weights and biases after the error correction operation, and $\textbf{W}_s$ denote the weights of the sparse model before correction. $b_d$, $\textbf{W}_d$, and $\textbf{X}_d$ are biases, weights, and input activation in the dense model, respectively. $f(\textbf{W, X})$ represents the convolutional operation performed by the layer on inputs $\textbf{X}$ with weights $\textbf{W}$.  $E$ and $\sigma$ are the mean and standard deviation operators, $\epsilon$ is a small constant.
In this way, we can correct the
error caused by the distribution shift of weights and biases.

\subsubsection{Reconstruction Input}
During the reconstruction procedure, we can either use the output of the previous reconstruction unit from the dense model as the input or opt for the output after the previous sparsified units. 
In current research, the choice of reconstruction inputs is not aligned, but we find that it greatly impacts the results. Since the absence of systematic investigation in previous work, we also benchmark this technique in our PTSBench.

\subsubsection{Reconstruction Granularity}
In addition to error correction and reconstruction input, we also benchmark the reconstruction granularity in our PTSBench. Specifically, 
the reconstruction process can be conducted at different granularities. 
Many post-training quantization methods~\cite{li2021brecq, wei2022qdrop, liu2023pd} prove that the reconstruction granularity has a large impact on quantization performance. 
However, the detailed impact of reconstruction granularities on a broad range of models is still unclear. Besides, the effectiveness of different reconstruction granularities still needs to be validated in PTS area. 
Therefore, inspired by \cite{li2021brecq}, we mainly benchmark three reconstruction granularities:
    
\begin{enumerate}
    \item \textbf{Single reconstruction.} Reconstruct the weights based on each individual layer, which represents the smallest reconstruction granularity.
    \item \textbf{Layer-wise reconstruction.} Reconstruct the weights at the layer level. For instance, in a CNN, reconstruct in a CONV-BN-ReLU combination pattern.
    \item \textbf{Block-wise reconstruction.} Reconstruct the weights based on the block level (e.g., residual block).
\end{enumerate}
In addition to the three aforementioned reconstruction granularities, some quantization methods also propose to use a net-wise reconstruction. However, although it is useful for quantization, we found this granularity will lead to poor performance because of overfitting. Hence, we do not include its performance for benchmarking in our PTSBench.



%% file: sec/overview.tex
\section{PTSBench: Tracks and Metrics}
\label{sec:overview}

\begin{figure*}[t]
  \centering
  \includegraphics[width=0.9\linewidth]{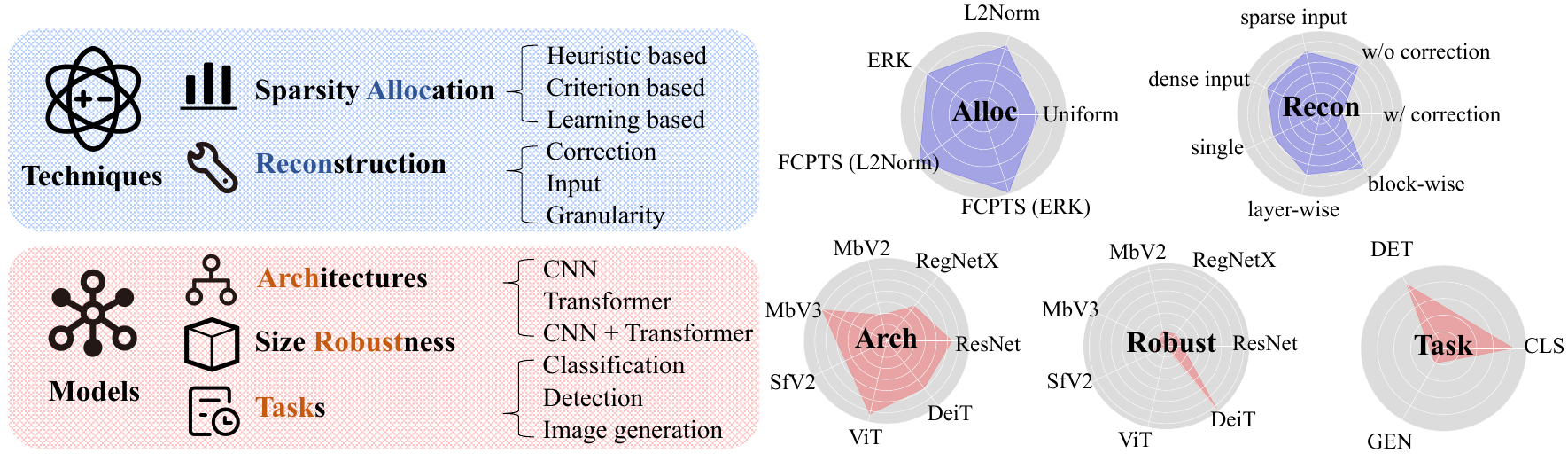 }
  \caption{Evaluation tracks of PTSBench. We benchmark the performance of PTS fine-grained algorithms and model sparsification abilities on a range of comprehensive evaluation tracks, including: "Sparsity Allocation", "Reconstruction", "Neural Architectures", "Model Size Robustness", and "Different Tasks". We illustrate an overview of the results of each track respectively on the right of the figure.}
  \label{fig:track}
\end{figure*}

\begin{table*}[t]
\small
\caption{Architecture repository.}
\label{tab:models}
\begin{tabular}{lll}
\toprule
Task                 & Arch. Family     & Archs                                                                                                              \\ \hline
\multirow{8}{*}{CLS} & ResNet \cite{he2016deep}          & ResNet-18, ResNet-32, ResNet-34, ResNet-50, ResNet-56, ResNet-101, ResNet-152                                      \\
                     & RegNetX   \cite{radosavovic2020designing}       & \begin{tabular}[c]{@{}l@{}}RegNetX-200M, RegNetX-400M, RegNetX-600M, RegNetX-800M, RegNetX-1600M, RegNetX-3200M, \\ RegNetX-4000M, RegNetX-6400M\end{tabular} \\
                     & MobileNetV2   \cite{sandler2018mobilenetv2}   & MobileNetV2-x0.5, MobileNetV2-x0.75, MobileNetV2-x1.0, MobileNetV2-x1.4                                            \\
                     & MobileNetV3   \cite{howard2019searching}   & MobileNetV3-x0.35, MobileNetV3-x0.5, MobileNetV3-x0.75, MobileNetV3-x1.0, MobileNetV3-x1.4                         \\
                     & ShuffleNetV2    \cite{ma2018shufflenet} & ShuffleNetV2-x0.5, ShuffleNetV2-x1.0, ShuffleNetV2-x1.5, ShuffleNetV2-x2.0                                         \\
                     & VGG \cite{simonyan2014very}             & VGG-19                                                                                                             \\
                     & ViT \cite{dosovitskiy2020image}             & ViT-B/16, ViT-B/32, ViT-L/16                                                                                       \\
                     & DeiT   \cite{touvron2021training}          & DeiT-Ti, DeiT-S, DeiT-B                                                                                            \\ \hline
\multirow{2}{*}{DET} & RetinaNet   \cite{lin2017focal}     & RetinaNet-R18, RetinaNet-R50                                                                                       \\
                     & SSD  \cite{liu2016ssd}            & MobileNetV1 SSD, MobileNetV2 SSD-Lite                                                                              \\ \hline
GEN                  & Stable Diffusion \cite{rombach2022high}& Stable Diffusion V2                                                                                                \\ \bottomrule
\end{tabular}
\end{table*}

This section presents PTSBench.
Our evaluation consists of 5 tracks and corresponding metrics, as shown in Fig.~\ref{fig:track}, which provide a comprehensive evaluation to address the limitations of current studies. Tab.~\ref{tab:models} presents the models we include in PTSBench.
A higher overall metric indicates better performance except for $OM_{robust}$ of the model size robustness track.

\subsection{Towards Fine-grained Algorithm}
\label{sec:overview:algo}

In our PTSBench, we benchmark "Sparsity Allocation" and "Reconstruction", which are two main procedures of PTS progress. To have a comprehensive evaluation, we conduct our experiments on 3 typical tasks: classification, detection, and image generation. 


\textbf{\textcircled{1} Sparsity Allocation.}In this track, we choose 4 sparsity allocation approaches for benchmarking. Uniform sparsity allocation is a widely adopted human heuristic-based method that allocates the same sparsity for each layer. L2Norm~\cite{han2015learning} and ERK~\cite{evci2020rigging} are both mask criterion-based methods, while the latter is more meticulously designed. We choose these two methods for their effectiveness and high citations. We also include FCPTS~\cite{gong2024fast} as a learning-based method, which is the only existing method.

To better quantify the performance, we define our overall metric (OM) by calculating the quadratic mean of the relative accuracies across 3 tasks as follows:
\begin{equation}
    \begin{aligned}
        OM_{alloc}=\sqrt{\frac{1}{3}\Bigg(\mathbb{E}^2\left(\frac{A^s_{CLS}}{A_{CLS}}\right)+\mathbb{E}^2\left(\frac{A^s_{DET}}{A_{DET}}\right)+\mathbb{E}^2\bigg(\frac{A_{GEN}}{A^s_{GEN}} \bigg) \Bigg)},
    \end{aligned}
\end{equation}
where $A_{*}$ and $A^s_{*}$ denotes the results obtained by calculating the metric of different tasks (i.e., accuracy for CLS, mAP for DET and FID~\cite{heusel2017gans} for GEN) of the dense and sparse models, respectively.
We take the average of different sparsity rates, architectures, and datasets in the mean operation $\mathbb{E}(\cdot)$. 
To remove the impact of reconstruction, we measure the accuracy of the model after sparsification without reconstruction. Noting that the Frechet Inception Distance (FID) is better with lower values, we take the reciprocal for the generation task to ensure the overall metric is in $[0,1]$ interval. 
We use the quadratic mean form to reduce the impact of outliers in the metric.

\textbf{\textcircled{2} Reconstruction.} In Section \ref{sec:reconstruction}, we introduce three fine-grained reconstruction techniques that may have a considerable impact on the effect while attracting little in-depth research. Therefore, we benchmark them in this track.
We define the performance gap with and without using a specific technique:
\begin{equation}
    \begin{aligned}
        R^s_*= A^{s,r}_*-A^s_*,
    \end{aligned}
\end{equation}
where $A^{s,r}_*$ denotes the results after the reconstruction.
For each fine-grained algorithm, we compute an overall metric. The overall metric of this track can be calculated by:
\begin{equation}
    \begin{aligned}
        OM_{recon}=\sqrt{\frac{1}{3}\Bigg(\mathbb{E}^2\left(\frac{R^s_{CLS}}{A_{CLS}}\right)+\mathbb{E}^2\left(\frac{R^s_{DET}}{A_{DET}}\right)+\mathbb{E}^2\bigg(exp\Big(\frac{A_{GEN}}{R^s_{GEN}}\Big) \bigg) \Bigg)}.
    \end{aligned}
\end{equation}
Here, we take the average of different sparsity rates, architectures, and datasets in the mean operation $\mathbb{E}(\cdot)$. As the FID score for the generation task has a different scale compared with the other two tasks, we take the exponential for the results of generation tasks.

\subsection{Towards Model Sparsification Ability}
\label{sec:overview:model}

We also benchmark the sparsity ability of models in PTSBench. Our evaluation includes 3 tracks: "Neural Architecture", "Model Size Robustness", and "Application Tasks".


\textbf{\textcircled{3} Neural Architecture.}
Although existing PTS methods evaluated the effectiveness on a wide range of models, the sparsity ability of the model itself remains uncovered. In our PTSBench, we benchmark the sparsity ability from the neural architecture aspect.
The overall metric is defined as: 
\begin{equation}
    \begin{aligned}
        OM_{arch} = \sqrt{\frac{1}{C}\sum^C_{i=1}\mathbb{E}^2\left(\frac{A^s_{size_i}}{A_{size_i}}\right)},
    \end{aligned}
\end{equation}
where $A_{size_i}$ denotes the set of accuracies for the $i$th size of a specific architecture (e.g., ResNet18 and ResNet50) under different sparsity rates. 
We take the average over different sparsity rates in $\mathbb{E}$.
$C$ denotes the number of model sizes. In simple terms, this metric evaluates the mean performance of models across different sparsity rates and model sizes within a specific architectural structure.

\textbf{\textcircled{4} Model Size Robustness.}
In real-world applications, the scale of deployed models varies due to different hardware resources. 
We also hope the PTS method will exhibit consistent effectiveness across models under the same architecture but with varying depths and widths. 
So, we benchmark model size robustness for PTS methods in this track.
The overall metric can be written as:
\begin{equation}
\label{eq:robust}
    \begin{aligned}
        OM_{robust} = std\left(\sqrt{\mathbb{E}^2\left(\frac{A^s_{size_i}}{A_{size_i}}\right)}\right),
    \end{aligned}
\end{equation}
where $A_{size_i}$ denotes the performance set for the $i$th size of a specific model type. $std(\cdot)$ denotes the standard deviation. We take the average overall sparsity rates in the mean operation.





\textbf{\textcircled{5} Application tasks. }
Models are often combined and augmented before being deployed in different applications. For example, ResNet is used as the backbone in detection tasks, with a neck and head connected afterward. In image generation tasks, CNNs and Transformers are combined for use. Therefore, there is a need to test sparsity ability from the aspect of tasks.
We benchmark three typical tasks in our PTSBench: classification, detection, and image generation. Similar to the overall metric for the neural architecture track, we build the overall metric for this track:
\begin{equation}
    \begin{aligned}
        OM_{task}=\sqrt{\frac{1}{N}\sum^N_{i=1}\mathbb{E}^2\left(\frac{A^s_{task_i}}{A_{task_i}}\right)},
    \end{aligned}
\end{equation}
where $A_{task_i}$ denotes the performance set from different models for the $i$th task. $N$ denotes the number of tasks.

%% file: sec/implementation.tex
\begin{table*}[]
\small
\centering
\tabcolsep=4px
\caption{Benchmarking the sparsity allocation strategy of PTS methods. Blue: best in a column. Light blue: second best in a column. Red: worst in a column. Light red: second worst in a column. }
\label{tab:alloc}
\begin{tabular}{lcccccccccccccccc}
\toprule
\multirow{2}{*}{Algorithms} & \multicolumn{5}{c}{CLS} & \multicolumn{5}{c}{DET} & \multicolumn{5}{c}{GEN} & \multicolumn{1}{c}{\multirow{2}{*}{$OM_{alloc}$}} \\ \cmidrule(r){2-6}\cmidrule(r){7-11}\cmidrule(r){12-16}
                            & 50 & 60 & 70 & 80 & $MS$ & 50 & 60 & 70 & 80 & $MS$ & 50 & 60 & 70 & 80 & $MS$ & \multicolumn{1}{c}{}                    \\ \midrule
Uniform                     & \cellcolor{red!25} 96.94   & \cellcolor{red!25} 84.52   & \cellcolor{red!25} 65.67  & \cellcolor{red!25} 32.55   &\cellcolor{red!25} 72.88    &  \cellcolor{red!25} 98.31  & \cellcolor{red!25} 93.75   &  \cellcolor{red!25} 80.89 &\cellcolor{red!25}  47.77  & \cellcolor{red!25} 80.94  & \cellcolor{red!25} 16.40& -   &  -  &  -  &  \cellcolor{red!25} 16.40 &    \cellcolor{red!25}                    63.59         \\
L2Norm                      &\cellcolor{blue!10}  98.22  &\cellcolor{white!100} 95.68   &\cellcolor{white!100} 86.17   &\cellcolor{red!10}  52.53   &\cellcolor{white!100}  84.05   & \cellcolor{red!10} 98.66   & \cellcolor{red!10} 96.92   & \cellcolor{red!10} 94.08   & \cellcolor{red!10}  65.86  &  \cellcolor{red!10} 88.97   & \cellcolor{white!100} 78.47 & -   & -   & -   &\cellcolor{white!100} 78.47    &\cellcolor{white!100}  83.94                                    \\
ERK                         &\cellcolor{white!100} 97.67   & \cellcolor{red!10} 94.34   & \cellcolor{red!10} 83.62   &\cellcolor{white!100}  56.13  & \cellcolor{red!10}  84.00   &  \cellcolor{blue!25} 99.23  &\cellcolor{white!100} 98.08   &\cellcolor{white!100} 95.14   & \cellcolor{white!100}  76.40 &\cellcolor{white!100} 92.30    &\cellcolor{red!10}  62.61   &  -  &  -  & -   &  \cellcolor{red!10}  62.61  &   \cellcolor{red!10} 80.61                                     \\
FCPTS (L2Norm)              & \cellcolor{blue!25} 98.35   & \cellcolor{blue!10}  96.64   &  \cellcolor{blue!10}  90.16  & \cellcolor{blue!10}  79.62   & \cellcolor{blue!10}   91.20   &\cellcolor{white!100}  99.00  & \cellcolor{blue!10}  98.25   &  \cellcolor{blue!10}  96.99  & \cellcolor{blue!10}  88.74   & \cellcolor{blue!10}  95.89    &  \cellcolor{blue!10} 87.80 &-    &  -  &  -  & \cellcolor{blue!10} 87.80    &  \cellcolor{blue!10}   91.69                                  \\
FCPTS (ERK)                 & \cellcolor{red!10} 97.51  & \cellcolor{blue!25}  97.85 & \cellcolor{blue!25} 93.79  &\cellcolor{blue!25} 88.57   & \cellcolor{blue!25} 94.55  & \cellcolor{blue!10}   99.19  &\cellcolor{blue!25} 98.58    & \cellcolor{blue!25} 97.16  & \cellcolor{blue!25} 91.95   &\cellcolor{blue!25} 96.76    &  \cellcolor{blue!25}  91.71  & -   & -   & -   & \cellcolor{blue!25}  91.71  & \cellcolor{blue!25} 94.36                                   \\ \bottomrule
\end{tabular}
\end{table*}

\section{Implementation Details}
\label{sec:implementation}




\textbf{Implementation details.} PTSBench is implemented using Nvidia A800 GPU. We follow the pipeline introduced in Sec.~\ref{sec:background:pipeline} to sparsify the dense model. In the reconstruction process, we use the SGD optimizer for optimization. The momentum is set as 0.9, and the learning rate is set as $1e-4$. We randomly select 1,024 images from the training datasets as our calibration datasets and calibrate for 20,000 epochs. The batch size is set as 64. 
We observe that when the sparsity rate is lower than 50\%, the performance drop after sparsification is negligible for almost all experiments. On the other hand, when the sparsity rate is higher than 80\%, almost all setups undergo a collapse in accuracy. Therefore, we calculate the overall metric based on $\{0.5, 0.6, 0.7, 0.8\}$ sparsity rates. For better presentation, we multiply all the $OM$ by 100.

\textbf{Evaluation settings.} 
For tracks 1 and 2, we conduct our experiments on three typical tasks: classification, detection, and image generation. For classification tasks, we evaluate VGG19\cite{simonyan2014very}, ResNet-18/32/50/56\cite{he2016deep}, RegNetX-200M/400M\cite{radosavovic2020designing}, MobileNetV2\cite{sandler2018mobilenetv2}, ViT\cite{dosovitskiy2020image} 
on ImageNet-1K~\cite{deng2009imagenet} and CIFAR-10/100~\cite{krizhevsky2009learning}. 
For detection task, we evaluate RetinaNet-r18/50~\cite{lin2017focal} on MSCOCO-2017~\cite{lin2014microsoft} and MobileNetV1-SSD, MobileNetV2-SSDLite~\cite{liu2016ssd} on PASCAL VOC07~\cite{everingham2010pascal}. For the image generation task, we evaluate Stable Diffusion~\cite{rombach2022high} on LSUN-Churches/Bedroom~\cite{yu2015lsun}. For tracks 3 and 4, we evaluate mainstream CNN and Transformer models. For CNN models, we include ResNet, RegNetX, MobileNetV2, MobileNetV3~\cite{howard2019searching}, and ShuffleNetV2~\cite{ma2018shufflenet}. For Transformer models, we include DeiT~\cite{touvron2021training} and ViT. All accuracies of models are measured on ImageNet-1K. For track 5,  we evaluate three tasks: classification, detection, and image generation.

%% file: sec/experiment.tex
\section{PTSBench Evaluation and Analysis}
\label{sec:experiment}

\subsection{Fine-grained Algorithm Tracks}

\subsubsection{Sparsity Allocation: A Well-allocated Sparsity Results In High Performance.}

We first present the evaluation results of different sparsity allocation strategies as shown in Tab.~\ref{tab:alloc}. To facilitate a more detailed analysis, we additionally report the root mean square components of the $OM_{alloc}$ for each task, denoted as MS, as well as its specific performance at each sparsity rate.

\textbf{The impact of sparsity allocation is crucial and significant.} Different sparsity allocation strategies vary greatly in results. Across various metrics, the Uniform strategy consistently shows the poorest performance, whereas learning-based methods uniformly exhibit good results. The gap between the two can reach up to 20\%. 
L2Norm behaves better than ERK, while FCPTS initiated with two strategies have different performance, which implies that initialization matters a lot for learning-based methods and ERK possesses better potential for fine-tuning than L2norm (i.e., ERK sparsity allocation is closer to an optimal distribution).

\begin{figure*}[t]
  \centering
  \includegraphics[width=0.9\linewidth]{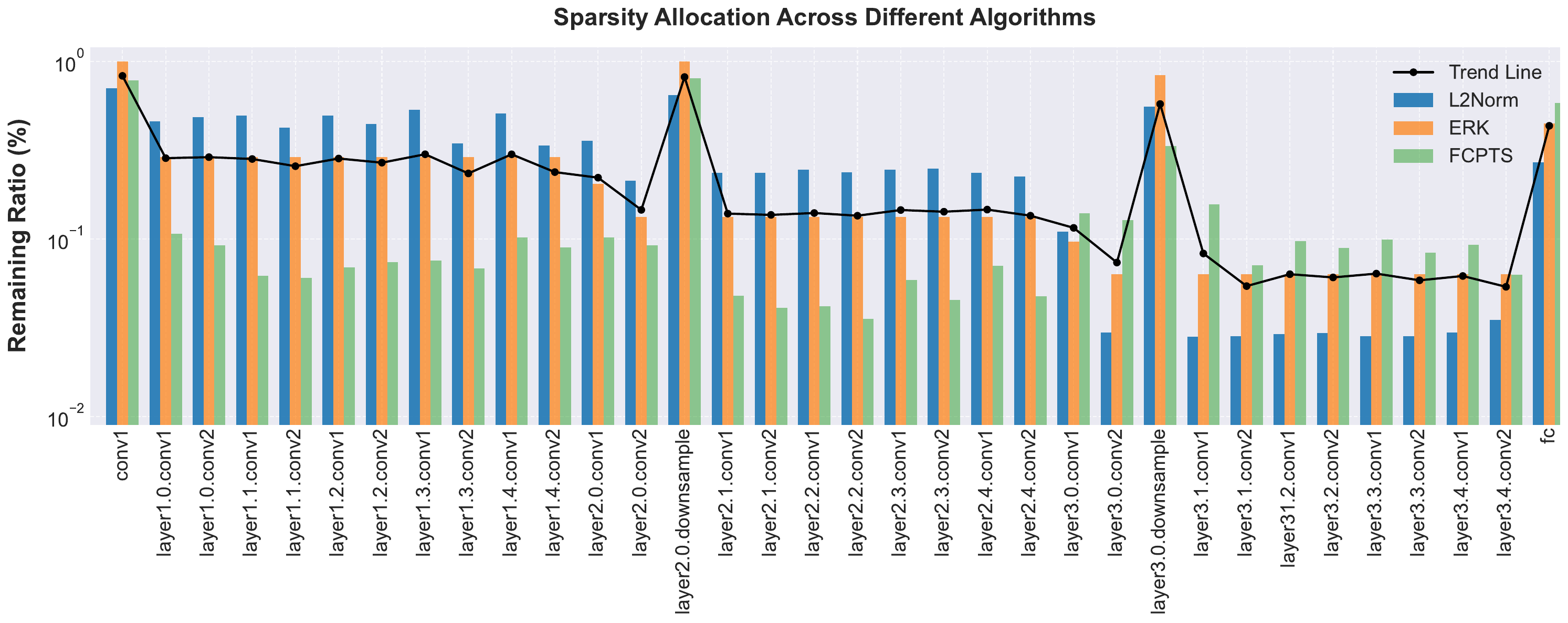 }
  \caption{Visualization of sparsity allocation of ResNet-32 at a sparsity rate of 90\% on CIFAR-100. }
  \label{fig:alloc}
\end{figure*}

\textbf{Effective sparsity rate allocation benefits from assigning lower sparsity rates to more sensitive layers.} 
We further visualize the sparsity allocation using different methods. Fig. \ref{fig:alloc} shows the sparsity allocation of ResNet-32 on CIFAR-100 datasets. We observe effective methods unanimously allocate a lower sparsity rate to the final layer. This is because the final layer is directly related to the network's output features, making the output highly sensitive to changes in the weights of the final layer. Thus, the final layer is unsuitable for large-scale sparsification. 


We also observe that ERK and L2Norm commonly allocate a relatively low sparsity rate for the downsample layers, which implies that these methods consider downsample layers as sensitive layers. On the other hand, FCPTS tends to remove more weight from these layers while achieving better performance. This indicates that the poor performance is caused by mistakenly preserving more weights for sparsity-friendly layers.




\begin{table}[]
\small
\caption{Comparison of the time consumption of different reconstruction granularities. The best results of each model are marked bolder.}
\label{tab:time_recon}
\resizebox{\linewidth}{!}{
\begin{tabular}{llcccc}
\toprule
\multirow{2}{*}{Model}            & \multirow{2}{*}{Granularity} & \multicolumn{4}{c}{Sparsity Rate (\%)} \\ \cmidrule{3-6} 
                                  &                              & 50     & 60     & 70     & 80     \\ \midrule
\multirow{3}{*}{ResNet-18}        & Single                       & 3h32m  & 3h34   & 3h31   & 3h28   \\
                                  & Layer-wise                   & 2h2m   & 1h58m  & 2h11m  & 2h5m   \\
                                  & Block-wise                   & \textbf{1h37m}  & \textbf{1h40m}  & \textbf{1h47m}  & \textbf{1h41m}  \\ \midrule
\multirow{3}{*}{ResNet-50}        & Single                       & 4h44m  & 4h40m  & 4h38m  & 4h31m  \\
                                  & Layer-wise                   & 3h30m  & 3h21m  & 3h8m   & 3h22m  \\
                                  & Block-wise                   & \textbf{2h7m}   & \textbf{2h12m}  & \textbf{2h12m}  & \textbf{2h10m}  \\ \midrule
\multirow{3}{*}{RegNet-200M}      & Single                       & 5h24m  & 5h30m  & 5h16m  & 5h22m  \\
                                  & Layer-wise                   & 1h10m  & 1h4m   & 1h8m   & 1h15m  \\
                                  & Block-wise                   & \textbf{31m55s} & \textbf{38m15s} & \textbf{40m22s} & \textbf{42m19s} \\ \midrule
\multirow{3}{*}{RegNet-400M}      & Single                       & 6h3m   & 6h2m   & 6h0m   & 6h0m   \\
                                  & Layer-wise                   & 1h33m  & 1h33m  & 1h48m  & 1h48m  \\
                                  & Block-wise                   & \textbf{1h1m}   & \textbf{1h9m}   & \textbf{1h19m}  & \textbf{1h19m}  \\ \midrule
\multirow{3}{*}{MobileNetV2-x1.0} & Single                       & 5h42m  & 5h33m  & 5h29m  & 5h39m  \\
                                  & Layer-wise                   & 42m43s & 35m0s  & 59m42s & 45m22s \\
                                  & Block-wise                   & \textbf{32m1s}  & \textbf{27m43s} & \textbf{44m5s}  & \textbf{44m4s}  \\ \bottomrule
\end{tabular}}
\vspace{-1em}
\end{table}

\subsubsection{Reconstruction:
Simple Adjustment Leads to Significant Sparsity Effect Difference.}

\begin{table*}[]
\small
\centering

\caption{Benchmarking the reconstruction techniques of PTS methods.}
\vspace{-0.5em}

\label{tab:recon}
\resizebox{\linewidth}{!}{
\begin{tabular}{lcccccccccccccccc}

\toprule
\multirow{2}{*}{Algorithms} & \multicolumn{5}{c}{CLS} & \multicolumn{5}{c}{DET} & \multicolumn{5}{c}{GEN} & \multirow{2}{*}{\textbf{$OM_{recon}$}} \\ \cmidrule(r){2-6}\cmidrule(r){7-11}\cmidrule(r){12-16}
                            & 50  & 60 & 70 & 80 & MS & 50  & 60 & 70 & 80 & MS & 50  & 60 & 70 & 80 & MS &                     \\ \midrule
w/ Correction               &   11.87   & 35.06  & 59.32&  32.90   &   36.68&   0  &   8.71  &  1.83  &   0  & 3.73 &88.32 &-    & -   & -   & 88.32  &     55.26            \\
w/o Correction              & 11.87    &  32.11   &   51.20 &  26.15   &   31.81  &   9.43  & 48.84   &  94.95  &  92.33  & 61.39   &  88.32& -   & -   &  -  &  88.32  &      64.76      \\ \midrule
Sparse Input                &  11.85  & 34.71  &59.07   &   33.86  & 35.87   &  9.28   &  48.69  & 94.79  & 92.02 &  61.20   & 88.32   &-    &   - &  -  & 88.32 &  65.40               \\
Dense Input                 &  11.30   &  33.91  &  56.44   &   23.40  &   32.84  &  8.73   &   47.46  &   92.40  & 88.40   &      59.25&  88.06 &   - &  -  &-    &   88.06   &  64.14              \\ \midrule
Singe                       &  9.40   &   33.26  &  59.73  &42.75  & 38.01 &  3.83   &  41.72   &  83.46 &71.89   & 50.24   &   88.11 & -   & -   &  -  & 88.11  & 62.54            \\
Layer-wise                  & 10.35  & 34.29 & 64.30  & 61.85  &  43.40  &  4.53  & 43.30 &88.85   &82.29 & 54.75  &88.18 &-    &   - &   - & 88.18&  64.95            \\
Block-wise                  &   10.68  & 35.77  & 67.14   &  69.21  &  46.32   &  8.73   & 47.26  &  92.15   &   85.54 & 58.43 &   88.32  &   - &    -&  -  & 88.32  & 66.76               \\ \bottomrule
\end{tabular}}
\end{table*}

\textbf{Error Correction behaves differently in different tasks.} 
We report results of this track in Tab.~\ref{tab:recon}. For classification tasks, we observe that error correction consistently results in high performance under different sparsity rates, which is similar to the experience in previous work\cite{lazarevich2021post}. However, the technique is performed diversely for detection and generation tasks. There is a significant collapse after applying error correction in detection tasks, while generation tasks seem to be insensitive toward the distortion of weights distribution. We hypothesize that
error correction on weight distribution may disrupt the location information learned by the model, which is critical for detection tasks. 

\textbf{Use sparse input is beneficial.}
From Tab.~\ref{tab:recon}, we find that in most settings, sparse input can reach higher performance than dense input, especially under a higher sparsity rate (e.g., for CLS tasks, 33.86 versus 23.40 under 80\% sparsity rate). This may be because using the output from the sparse model can make the PTS algorithm aware of the reconstruction error from the previous layer, which avoids error accumulation across the network.

\textbf{Block-wise reconstruction is always the best.} Block-wise reconstruction achieves the best results under most configurations, and layer-wise reconstruction outperforms single reconstruction. For example, when sparsifying classification models at an 80\% sparsity rate, block-wise reconstruction can outperform layer-wise by up to 3\% and surpass single reconstruction by 16\%. Block-wise reconstruction also has the advantage of time consumption (see Tab.~\ref{tab:time_recon}). Therefore, it is beneficial to use block-wise reconstruction in post-training sparsity.
\begin{table}[t]
\small
\caption{Model sparsity potential and model size robustness. Blue and red: best and worst in a column. Light blue and light red: second best and second worst in a column.}
\label{tab:arch}
\tabcolsep=3px
\begin{tabular}{lcccccc}
\toprule
\multirow{2}{*}{Models} & \multicolumn{4}{c}{Sparsity Rate (\%)} & \multirow{2}{*}{$\textbf{OM}_{\textbf{arch}}$} & \multirow{2}{*}{$\textbf{OM}_{\textbf{robust}}$}\\ \cmidrule{2-5}
                        & 50       & 60      & 70      & 80      &                     \\ \midrule
ResNet                  &  \cellcolor{blue!10}   99.38     &\cellcolor{blue!25}  97.86      &  \cellcolor{blue!25} 91.38     &  \cellcolor{white!100}    48.36     &  \cellcolor{white!100}    86.81   &  \cellcolor{red!10}   0.286         \\
RegNetX                 &  \cellcolor{white!100}    97.98      & \cellcolor{white!100}     94.57     &  \cellcolor{white!100}      83.65  &    \cellcolor{red!10}   45.17  &  \cellcolor{red!10}  83.04   &    \cellcolor{white!100}     0.270       \\
MobileNetV2             &   \cellcolor{red!10}   97.32      &   \cellcolor{red!10}  93.48    &     \cellcolor{red!25} 79.80 & \cellcolor{red!25}   29.51    &   \cellcolor{red!25}   79.77   &    \cellcolor{white!100}   0.219         \\
MobileNetV3             & \cellcolor{white!100}   97.93    & \cellcolor{white!100}    95.54    & \cellcolor{white!100}     89.14     &   \cellcolor{blue!25} 65.32     &   \cellcolor{blue!10} 87.94   &  \cellcolor{blue!10}    0.172        \\
ShuffleNetV2            &  \cellcolor{red!25} 96.50       &   \cellcolor{red!25} 92.77     &     \cellcolor{red!10}  82.96    &    \cellcolor{white!100}   54.04    &   \cellcolor{white!100}    83.25  &    \cellcolor{white!100}      0.258      \\ \midrule
ViT                     &  \cellcolor{blue!25}  99.50      &   \cellcolor{blue!10} 97.52      &   \cellcolor{blue!10} 89.99     & \cellcolor{blue!10}  62.41      &   \cellcolor{blue!25} 88.61   &  \cellcolor{blue!25}   0.014         \\
DeiT                    & \cellcolor{white!100}  98.08      &  \cellcolor{white!100}  95.53       &  \cellcolor{white!100}     85.81    & \cellcolor{white!100}    58.25     & \cellcolor{white!100}  85.88     &      \cellcolor{red!25}   1.485    \\ \bottomrule
\end{tabular}
\vspace{-1em}
\end{table}

\subsection{Model Tracks}


\subsubsection{Neural Architecture: Sparsity Potential Varies Across Different Architectures.} 

\textbf{Models based on attention mechanisms possess greater sparsity potential.} 
We report the evaluation results under different sparsity rates and overall metrics in Tab.~\ref{tab:arch}. We can find that ViT, MobileNetV3, and DeiT have their Overall Metrics (OM) positioned among the top across all models evaluated (top 1, 2, and 4, respectively). Compared to MobileNetV2, which possesses the worst $OM_{arch}$,  MobileNetV3 significantly has its sparsity potential enhanced. The main difference between these two similar architectures is that MobileNetV3 introduced a Squeeze-and-Excitation Block~\cite{howard2019searching}, which is considered a lightweight attention mechanism\cite{hu2018squeeze}. ViT and DeiT are also based on attention structure. 
The attention mechanism allows the network to put more information on neurons
that are crucial for the final task, and reduce the information on less important neurons.
Naturally, in scenarios of sparsification, it enables more effective preservation of information critical to performance.
Therefore, we can choose an attention-based structure if we need to sparsify the model for further deployment.

\textbf{Training strategy can impact its sparsity potential.} 
DeiT has nearly the same architecture compared to ViT, while has poorer performance than ViT. This may be because of significant differences in their training strategies. ViT is pre-trained on extremely large datasets, such as JFT-300M, which likely enables it to learn more general feature representations. This makes it more robust against sparsification. 
While DeiT employs knowledge distillation as one of its core strategies, which aids in training efficient models with less data. Thus, we can utilize large-scale pre-training to obtain a sparsity-friendly model.
\begin{figure}[t]
  \centering
  \includegraphics[width=0.9\linewidth]{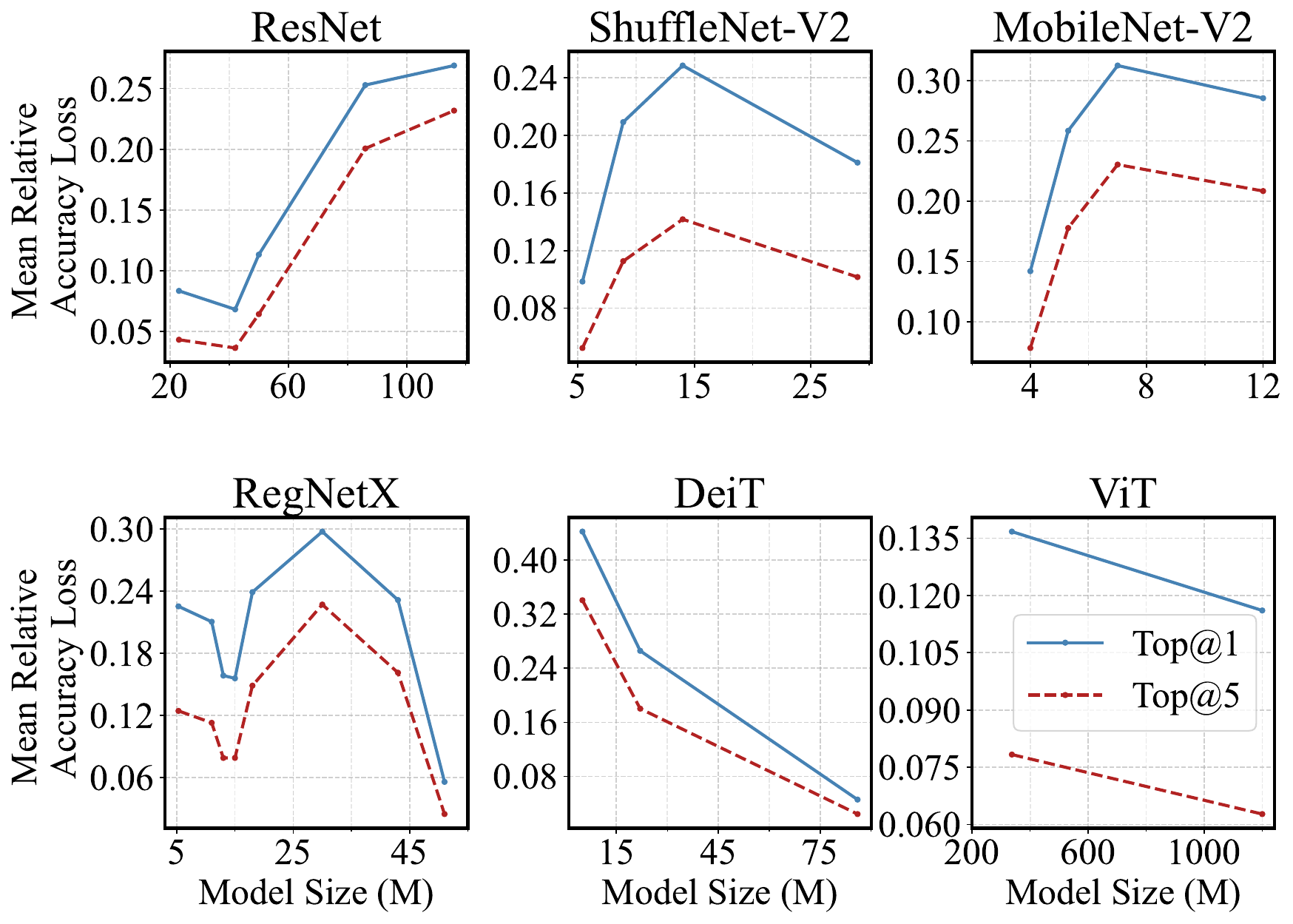 }
  \caption{Mean relative accuracy loss of different model sizes.}
  \vspace{-1em}
  \label{fig:size}
\end{figure}

\subsubsection{Model Size Robustness: Different Model Architectures Tends To Vary In Model Size Robustness.}

\textbf{High sparsity potential for specific model $\neq$ high sparsity potential for this model type.} 
The results of evaluated models are shown in Tab.~\ref{tab:arch}. Interestingly, we observe that DeiT and ResNet, which perform well on the $OM_{arch}$, exhibit poor performance on $OM_{robust}$, while ViT and MobileNetV3 behave consistently on both metrics. 
This suggests that the sparsity potential of a model and its size robustness don't exhibit a strict positive correlation. However, models with high sparsity potential are considered more likely to possess good model size robustness. 

\textbf{DeiT fails to achieve good model size robustness.}
DeiT is notably inferior to other models in terms of model size robustness. We hypothesize that DeiT is trained using knowledge distillation with less data. So the information may vary in different model sizes. Therefore, it is less robust to different model sizes.

\textbf{A larger model size does not necessarily mean better sparsity ability.} 
In our PTSBench, a counter-intuitive finding is large model size cannot guarantee better sparsity ability. 
To demonstrate this finding, we present the mean relative accuracy drop after model sparsification of each model size, which is defined as the average performance drop over the performance of the dense model under different sparsity rates. In Fig.~\ref{fig:size}, we can draw the following discoveries.
(1) For CNNs, the mean relative accuracy loss first increases and then decreases as the model size increases. So, not to choose a middle-sized network is the key point for CNN-based network architectures.
(2) For DeiT and ViT, the mean relative accuracy loss decreases as the model size increases, which shows that the Transformer is more amenable to sparsification under a larger model size. 


\begin{table}[t!]
\small
\caption{Benchmarking sparsification potential of different tasks. Blue: best in a column.  Red: worst in a column.}
\label{tab:tasks}
\begin{tabular}{lccccc}
\toprule
\multirow{2}{*}{Tasks} & \multicolumn{4}{c}{Sparsity Rate (\%)}                                                            & \multirow{2}{*}{$\textbf{OM}_{\textbf{task}}$} \\ \cmidrule{2-5}
                       & \multicolumn{1}{c}{50} & \multicolumn{1}{c}{60} & \multicolumn{1}{c}{70} & \multicolumn{1}{c}{80} &                     \\ \midrule
CLS                    & \cellcolor{white!100}    98.24                   &       \cellcolor{white!100}       95.68            &     \cellcolor{white!100}      86.17               &        \cellcolor{white!100}      52.53            &    \cellcolor{white!100}        84.05           \\ 
DET                    &    \cellcolor{blue!10}    98.66                  &    \cellcolor{blue!10}         96.92             &   \cellcolor{blue!10}        94.08               &    \cellcolor{blue!10}       65.86               &    \cellcolor{blue!25}      88.97            \\ 

GEN           &     \cellcolor{red!10}       78.06   &   \cellcolor{red!10}   5.44    &     \cellcolor{red!10}   0.18        &    \cellcolor{red!10}      0      &  \cellcolor{red!25} 21.02 \\ \bottomrule
\end{tabular}
\vspace{-1em}
\end{table}

\begin{table*}[t!]
    \centering
    \small
    \vspace{-0.5em}
    \begin{tcolorbox}
    \begin{tabular}{p{0.97\columnwidth} c}
    \vspace{-1em}
      \VarSty{ \bf Track 1: Sparsity Allocation }   & \hspace{-5.3cm} \normalsize\bf PTS Fine-grained Techniques \\
        \begin{itemize}[wide, labelindent=0pt]
        \setlength{\itemsep}{0pt}
        \setlength{\parsep}{0pt}
        \setlength{\parskip}{0pt}
        \vspace{-1em}
        \itshape
            \item Priority: learning-based \textgreater criterion-based \textgreater heuristic-based
            \item For learning-based method, 
            choosing appropriate initialization for sparsity allocation can boost performance.
            \item The final layer is unsuitable for a high sparsity rate. We recommend keeping it dense for better performance.
        \vspace{-1em}
        \end{itemize}
         & \\
          \VarSty{\bf Track 2: Reconstruction}&\\
          \begin{itemize}[wide, labelindent=0pt]
        \setlength{\itemsep}{0pt}
        \setlength{\parsep}{0pt}
        \setlength{\parskip}{0pt}
          \vspace{-1em}
        \itshape
            \item Error correction: it is effective for classification, but not for detection and image generation.
            \item Reconstruction input: sparse input is better than dense input. In other words, it is better to use the output from the last reconstruction unit as the input.
            \item Reconstruction granularity: block-wise \textgreater layer-wise \textgreater single.
            \vspace{-1.5em}
        \end{itemize} & \\
         \hrulefill & \\
         \VarSty{\bf Track 3: Neural Architecture}&\hspace{-4.3cm} \normalsize \bf Model Sparsity Ability  \\
         \begin{itemize}[wide, labelindent=0pt]
        \setlength{\itemsep}{0pt}
        \setlength{\parsep}{0pt}
        \setlength{\parskip}{0pt}
          \vspace{-0.8em}
            \itshape
            \item Sparsity potential: ViT \textgreater MbV3 \textgreater ResNet \textgreater Deit \textgreater SfV2 \textgreater RegNetX \textgreater MbV2.
            \item Structure: The attention mechanism is a more sparsity-friendly model structure.
            \item Training strategy: large-scale datasets can enhance the model sparsity ability.
            \vspace{-1em}
        \end{itemize} & \\
        \VarSty{\bf Track 4: Model Size Robustness}&\\
        \begin{itemize}[wide, labelindent=0pt]
        \setlength{\itemsep}{0pt}
        \setlength{\parsep}{0pt}
        \setlength{\parskip}{0pt}
          \vspace{-0.9em}
        \itshape
            \item Robustness: ViT \textgreater MbV3 \textgreater MbV2 \textgreater SfV2 \textgreater RegNetX \textgreater ResNet \textgreater DeiT.
            \item Transformer-based models are more robust against different scales.
            \vspace{-1em }
        \end{itemize} & \\
        \VarSty{\bf Track 5: Application Tasks}&\\
        \begin{itemize}[wide, labelindent=0pt]
        \setlength{\itemsep}{0pt}
        \setlength{\parsep}{0pt}
        \setlength{\parskip}{0pt}
          \vspace{-0.85em}
        \itshape
            \item Performance: detection \textgreater classification \textgreater image generation. 
            \item PTS method still needs further development for image generation.
            \vspace{-1.5em}
        \end{itemize} & \\
        
    \end{tabular}
    
    \end{tcolorbox}
    \caption{Takeaway conclusions of our PTSBench.}
    \label{tab:tech}
    \vspace{-2.5em}
\end{table*}

\subsubsection{Application Tasks: The PTS Method Needs Further Development In Generation Tasks.}

\textbf{Detection has better performance than classification.} 
 We present the benchmark results of different tasks in Tab.~\ref{tab:tasks}. From Tab. \ref{tab:tasks}, we find that $OM_{task}$ can reach up to 88.97 in detection tasks, whereas classification task scores 84.05. This implies that attaching subsequent structures (such as the neck \cite{lin2017feature} and head parts in detection models) to a backbone does not reduce sparsity potential; it can even make the model more sparsity-friendly due to the introduction of additional parameters.

\textbf{Image generation task requires specific designed PTS.} 
Generation tasks can only maintain precision at a 50\% sparsity rate, with a collapsing performance on higher sparsity rates. This may be because these methods do not consider the unique operating mechanisms of diffusion models and the distribution of their parameters. This implies the PTS methods specifically designed for image generation are required.

\subsection{Overall Results}

\begin{figure}[t]
  \centering
  \includegraphics[width=0.8\linewidth]{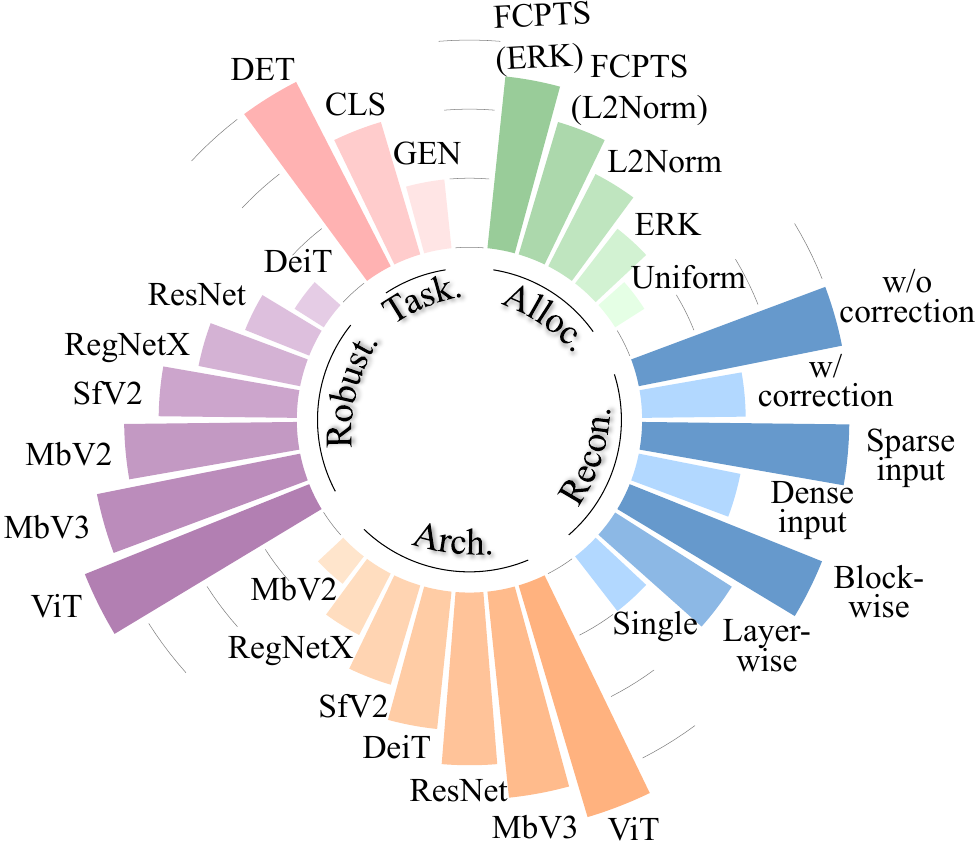 }
  \vspace{-1em}
  \caption{The overall evaluation results of PTSBench.}
  \label{fig:discussion}
  \vspace{-1.5em}
\end{figure}

Through extensive experiments and evaluations, we reach several take-away conclusions.
The overall evaluation results are shown in Fig.~\ref{fig:discussion}, and the results in different colors are ranked from best to worst. 
We also summarize the following valuable and effective takeaway conclusions on both fine-grained techniques and model sparisy ability perpectives as shown in Tab.~\ref{tab:tech}. These conclusions can help further works design more effective and efficient PTS algorithms and make contributions to practical PTS applications.

%% file: sec/discussion.tex
\section{Conclusion}
\label{sec:discussion}

In this paper, we systematically propose a \textbf{P}ost-\textbf{T}raining \textbf{S}parsification \textbf{Bench}mark called \textbf{PTSBench}, which is the first comprehensive benchmark towards the post-training sparsity (PTS). From an algorithm perspective, we benchmark 10+ PTS components on 3 computer vision tasks. From a model perspective, we benchmark 40+ network architectures. PTSBench aims to establish a comprehensive and in-depth analysis of PTS algorithms, providing useful technical guidance for future research. Our benchmark is highlighted by fertilizing the community by providing the following: 
(a) comprehensively evaluate models from the perspective of PTS.
(b) new observations towards a better understanding of the PTS fine-grained algorithms. 
(c) an open-source platform for systematically evaluating the model sparsification ability and pluggable sparsification algorithms. We plan to explore a broader range of model architectures and tasks in future work. We hope our PTSBench can provide useful advice for future studies.

Our PTSBench also has limitations: (1) We benchmark PTS methods on three vision tasks, and it is better to include more tasks like natural language processing in our PTSBench. (2) The number of PTS algorithms available for study is relatively small in the current research. As the PTS community becomes more fertilized, it is desirable also to include more PTS algorithms in our PTSBench. Considering the aforementioned limitations, we will continue to include more methods and tasks in our PTSBench platform.